\documentclass[runningheads,a4paper]{llncs}
\usepackage{todonotes}
\usepackage{rotating}
\usepackage{epstopdf}
\usepackage[british]{babel}
\usepackage[T1]{fontenc}
\usepackage[utf8]{inputenc}
\PassOptionsToPackage{hyphens}{url}
\usepackage[breaklinks,hidelinks]{hyperref}
\usepackage{breakurl}
\usepackage{tikz}
\usepackage{amssymb}
\usepackage{multirow}
\usepackage{paralist}
\usepackage{floatrow}
\usepackage{tikz}

\begin{document}
\sloppy
\mainmatter 

\title{A ROS Multi-Ontology References Service: OWL Reasoners and Application Prototyping Issues\footnote{Published in the CEUR-WS Proceedings of the 5th Italian Workshop on Artificial Intelligence and Robotics A workshop of the XVII International Conference of the Italian Association for Artificial Intelligence (AI*IA), 22-23 November 2018, Trento, Italy, Vol-2352, pages 36-41.}}
\titlerunning{A ROS Multi Ontology References (ARMOR) Service} 
\toctitle{A ROS Multi-Ontology References Service: OWL Reasoners and Application Prototyping Issues}

\author{Luca Buoncompagni\textsuperscript{\dag} \and Alessio Capitanelli\textsuperscript{\dag} \and Fulvio Mastrogiovanni}
\authorrunning{Buoncompagni et al.}
\tocauthor{Luca Buoncompagni, Alessio Capitanelli, and Fulvio Mastrogiovanni}

\institute{All the authors are affiliated with the Department of Informatics, Bioengineering, Robotics and Systems Engineering, University of Genoa, Via Opera Pia 13, 16145, Genoa, Italy.
Corresponding authors’ e-mails: \href{mailto:luca.buoncompagni@edu.unige.it}{luca.buoncompagni@edu.unige.it}.
\textsuperscript{\dag}These authors contributed equally to this work.
}
\maketitle

\begin{abstract}
This paper introduces a ROS Multi Ontology References (ARMOR) service, a general-purpose and scalable interface between robot architectures and OWL reasoners.
ARMOR addresses synchronization and communication issues among heterogeneous and distributed software components.
As a guiding scenario, we consider a prototyping approach for the use of symbolic reasoning in human-robot interaction applications.
\end{abstract}

\section{Introduction}
The challenge of sharing and communicating information is crucial in complex human-robot interaction (HRI) scenarios.
Ontologies and symbolic reasoning are the state of the art approach for a natural representation of knowledge, especially within the Semantic Web domain, and it has been adopted to achieve high expressiveness \cite{DL}.
Since symbolic reasoning is a high complexity problem, optimizing its performance requires a careful design of the knowledge \emph{resolution}.
Specifically, a robot architecture requires the integration of several components implementing different behaviors and generating a series of beliefs.
Most of the components are expected to access, manipulate, and reason upon a run-time generated representation of knowledge grounding robot behaviors and perceptions through formal axioms, with soft real-time requirements.
     
The Robot Operating System (ROS) 
is a \textit{de facto} standard for robot software development, which allows for modular and scalable robot architecture designs.
Currently, some approaches exist to integrate a semantic representation in ROS, such as the KnowRob \cite{tenorth2009knowrob} which provide a complete framework of relevant ontologies, or the native ROS support of MongoDB\footnote{\url{https://www.mongodb.com}.}, which can also be used to provide a suitable representation for semantic querying.
Unfortunately, none of these supports the study of advanced reasoning paradigms, and they heavily rely on \textit{ad hoc} reasoning solutions, significantly limiting their scope.
We argue that this fact affects the study of different approaches to semantics in Robotics. 
For instance, it limits our capability to explore novel semantic representations of perceptions, which offers similar but \textit{not equivalent} beliefs. 
We lack a standardized general framework to work with ontologies, natively supporting symbolic logic and advanced reasoning paradigms.

The Ontology Web Language (OWL)\footnote{\url{https://www.w3.org/TR/owl-guide}.} is a standard representation supporting several reasoning interfaces, e.g., Pellet \cite{pellet}, and logic formalisms, e.g., the Allen's Algebra \cite{allen}.
Thus, it can be a solid foundation for a framework for symbolic reasoning in Robotics.
OWL is based on the separation between terminological and assertional knowledge, referred to as different \emph{boxes} (Tbox and Abox).
Typically, in Robotics scenarios, we design a \emph{static} semantics for the beliefs to be represented in the TBox.
Then, we populate the ABox through \emph{individuals} defined using \emph{types} and \emph{properties} and, at run-time, we classify knowledge using \emph{instance checking}.
We argue that, due to the high complexity of HRI scenarios, the possibility of a dynamic semantics in the TBox is desirable as well.
For instance, it could be used to \emph{learn} new types for classification.
This leads us to a study requiring reasoning heuristics to be compared, components to be shared, and different semantics to be adapted. 
     
For this purpose, we propose the ROS Multi Ontology References\footnote{\url{https://github.com/EmaroLab/ARMOR}.} (ARMOR).
ARMOR is an open source \emph{service} which manipulates and queries multiple OWL ontologies.
It provides access to a set of dynamic ontologies, handling also the synchronizations among different components in the architecture.
Therefore, it is a convenient tool for managing knowledge representation supported by advanced reasoners.

\section{System's Architecture}
     
\begin{figure}[!t]
    \centering
    \resizebox{.9\textwidth}{!}{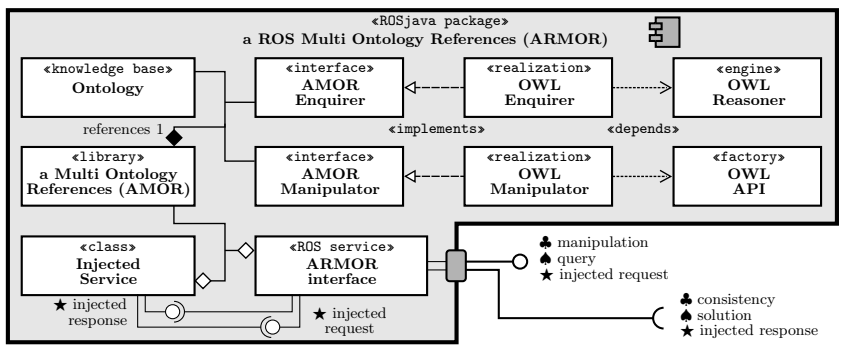}
    \caption{The UML diagram of ARMOR. It accepts two types of request and respond accordingly ({$\clubsuit,\spadesuit$}). The request and response {$\bigstar$} are shown as an example where the Injected Service extends the services of ARMOR using AMOR.\vspace{-.8em}}
    \label{fig:arch}
\end{figure}

Figure \ref{fig:arch} shows a schematic representation of ARMOR.
It interfaces the OWL API \cite{OWL} and reasoners through the Java-based Multi Ontology References library (AMOR). Then, ARMOR exposes AMOR functionalities as a service to ROS-based architectures, relying on the support for Java in ROS (ROSJava\footnote{\url{https://github.com/rosjava}.}).
ARMOR messages have been designed to accommodate all OWL functionalities.
Nevertheless, we have implemented only an exhaustive subset of those features so far (i.e., only common \textit{run-time} operations).
Indeed, ontology managers are not distributed across satellite components of a ROS architecture.
Instead, dedicated components are in charge of management, while others only provide knowledge axioms, possibly at run-time.
With ARMOR, it is possible to \emph{inject} in the service procedure managing symbols from a centralized perspective, based on the functionalities provided by AMOR.
Nevertheless, complex static representations can always be defined also off-line with dedicated software, e.g., Prot\'eg\'e\footnote{\url{http://protege.stanford.edu}.}. 

\subsection{The ARMOR Core: AMOR}
The core library, referred to as AMOR, contains a \emph{map} of instantiated ontologies, where the \emph{key} is a unique name identifier used to access the corresponding ontology in a thread-safe manner.
AMOR provides several common operations, such as those to create individuals, assign their properties, make them disjointed and classify them, to name a few.
Furthermore, AMOR ensures complete accessibility to the OWL API\footnote{\url{http://owlapi.sourceforge.net}.} features for compatibility and extendability purposes.
For example, AMOR allows for invoking reasoners by specifying their \texttt{OWLReasoner} \emph{factory}, i.e., the unique package of its Java implementation, which assures compatibility with all OWL reasoners.

In the current implementation, we interface several properties that are useful to tune the AMOR behaviour, e.g., the buffering of the manipulations or a continuous reasoner update, using the standard ROS Parameter Server, as well as parameters for debugging purposes such as toggling a Graphical User Interface (GUI) for visualising ontology states on-line.

\vspace{-.2em}
\subsection{The ARMOR Interface}
The ARMOR interface is based on a ROS message structure (i.e., a triple) for the use of the AMOR functionalities from any node in the architecture, even when the development language is different from Java (e.g., Python\footnote{An ARMOR client is available at \url{https://github.com/EmaroLab/ARMOR_py_api}.}
and C++ are the most common languages in Robotics development).
Such a message is composed of: 
\begin{inparaenum}
     \item the \emph{client name}, which is used by the service to identify different callers,
     \item the \emph{reference name}, indicating the operation's target reference, and
     \item the \emph{command} to execute, i.e., \texttt{add}, \texttt{remove}, \texttt{replace}, \texttt{query}, \texttt{load}, \texttt{mount}, etc.\footnote{Listed at \url{https://github.com/EmaroLab/ARMOR/blob/master/commands.md}.}
Each of those commands may be further refined by:
\begin{inparaenum}
\item the \emph{primary} and \emph{secondary specifiers}, which augment command labels, e.g., \texttt{add(\emph{individual}, \emph{class})} or \texttt{remove(\emph{individual}, \emph{property})}, and
\item the \emph{arguments}, a list of entities in the reference parameterising the command, e.g., $\langle$\texttt{add(\emph{class}) \textquotedbl Sphere\textquotedbl}$\rangle$, or even $\langle$\texttt{add(\emph{property},\emph{individual}) \textquotedbl hasNorth\textquotedbl} \texttt{\textquotedbl LivingRoom\textquotedbl} \texttt{\textquotedbl Corridor\textquotedbl}$\rangle$. 
\end{inparaenum}
\end{inparaenum}
An ARMOR call is based on one or more messages with the same structure.
When such a request is sent, the service manipulates or queries the ontology with the given directives.
Then, it returns whether the ontology is consistent, eventual error codes with their description, and the names of the queried entities, if requested. 
In other words, the interfaces {\footnotesize$\clubsuit$} and {\footnotesize$\spadesuit$} in Figure \ref{fig:arch}, and {\footnotesize$\bigstar$} can be defined for each specific injected service, if any.

One advanced feature of ARMOR is the possibility of flexibly synchronising all operations.
This follows a \emph{mounting}/\emph{unmounting} paradigm, where one or more nodes identified by the same \textit{client name} can prevent other nodes from manipulating a given ontology, in order to ensure manipulation consistency. 
On the contrary, queries are always allowed, except during reasoning time.
Calls to busy ontologies will report a mounting issue, and the user can choose how to handle this situation.

\vspace{-.3em}
\section{Applications and Conclusions}
We are currently using ARMOR in different applications, but here we mention only two of them.
The first is aimed at implementing a dynamic PDDL problem generator.
This approach uses descriptions of the \emph{predicates} and \emph{objects} in a tabletop scenario to infer unsatisfied norms and consequently generate goals \cite{alessioTh}.
The system has been integrated with ROSPlan\footnote{\url{https://github.com/KCL-Planning/ROSPlan}.} by substituting the internal semantic data structure with ARMOR and a suitable ontology. 

The second application is a system to learn by demonstrations the arrangement of objects in the robot's workspace by mapping their properties into the TBox.
In particular, we used an injected service\footnote{\url{https://github.com/EmaroLab/injected_armor_pkgs}.} in ARMOR for performing scene learning and classification in a scenario where a robot explains its beliefs to a human, which might want to correct it through dialogues \cite{architecturalSIT}.

This paper introduces the ARMOR service to manipulate OWL ontologies and query their reasoners in a ROS-based architecture.
ARMOR services are available through a flexible message allowing for the direct access OWL features from any component of the architecture.
It ensures synchronisation between client calls and flexibility through procedure injection.
Also, ARMOR allows for an easy interface between robotic architectures and OWL representations, and we practically showed it during a tutorial presented at the ROS Development Conference 2018\footnote{\url{http://www.theconstructsim.com/ros-developers-online-conference-2018-rdc-worldwide/ros-developers-conference-speaker-alessio-capitanelli}. Code available at \url{https://github.com/EmaroLab/armor_rds_tutorial}.}.
The tutorial is focused on the control of mobile robots based on a topological environment representation and SLAM.

\bibliography{bib}

\begin{thebibliography}{1}
\providecommand{\url}[1]{\texttt{#1}}
\providecommand{\urlprefix}{URL }

\bibitem{allen}
Allen, J.F.: Maintaining knowledge about temporal intervals. Communications of
  the ACM  26(11),  832--843 (1983)

\bibitem{DL}
Baader, F., Horrocks, I., Sattler, U.: Description logics as ontology languages
  for the semantic web. In: Mechanizing Mathematical Reasoning, pp. 228--248.
  Springer (2005)

\bibitem{architecturalSIT}
Buoncompagni, L., Mastrogiovanni, F.: Dialogue-based supervision and
  explanation of robot spatial beliefs: a software architecture perspective.
  In: 2018 27th IEEE International Symposium on Robot and Human Interactive
  Communication (RO-MAN). pp. 977--984 (Aug 2018)

\bibitem{alessioTh}
Capitanelli, A., Mastrogiovanni, F.: An ontology-based hybrid architecture for
  planning and robust execution in tabletop scenarios. In: Proceedings of the
  3rd Italian Workshop on Artificial Intelligence and Robotics A workshop of
  the XV International Conference of the Italian Association for Artificial
  Intelligence (AI*IA 2016). vol. 1834, pp. 31--35. CEUR-WS, Genova, Italy (nov
  2016)

\bibitem{OWL}
Horridge, M., Bechhofer, S.: The owl api: A java api for owl ontologies.
  Semant. web  2(1),  11--21 (Jan 2011)

\bibitem{pellet}
Sirin, E., Parsia, B., Grau, B.C., Kalyanpur, A., Katz, Y.: Pellet: A practical
  owl-dl reasoner. Web Semantics: science, services and agents on the World
  Wide Web  5(2),  51--53 (2007)

\bibitem{tenorth2009knowrob}
Tenorth, M., Beetz, M.: Knowrob -- knowledge processing for autonomous personal
  robots. In: 2009 IEEE/RSJ International Conference on Intelligent Robots and
  Systems. pp. 4261--4266. IEEE (Oct 2009)

\end{thebibliography}

\end{document}